# Investigation of Enhanced Inertial Navigation Algorithms by Functional Iteration

**Hongyan Jiang**

**Maoran Zhu**, Member, IEEE

**Yuanxin Wu**, Senior Member, IEEE

Shanghai Jiao Tong University, Shanghai, China

*Abstract*—The defects of the traditional strapdown inertial navigation algorithms become well acknowledged and the corresponding enhanced algorithms have been quite recently proposed trying to mitigate both theoretical and algorithmic defects. In this paper, the analytical accuracy evaluation of both the traditional algorithms and the enhanced algorithms is investigated, against the true reference for the first time enabled by the functional iteration approach having provable convergence. The analyses by the help of MATLAB Symbolic Toolbox show that the resultant error orders of all algorithms under investigation are consistent with those in the existing literatures, and the enhanced attitude algorithm notably reduces error orders of the traditional counterpart, while the impact of the enhanced velocity algorithm on error order reduction is insignificant. Simulation results agree with analyses that the superiority of the enhanced algorithm over the traditional one in the body-frame attitude computation scenario diminishes significantly in the entire inertial navigation computation scenario, while the functional iteration approach possesses significant accuracy superiority even under sustained lowly dynamic conditions.

This work was supported in part by National Natural Science Foundation of China (62273228), Postdoctoral Science Foundation and Fellowship Program of CPSF (GZC20231591, 2023M732227), and Shanghai Jiao Tong University Scientific and Technological Innovation Funds *(Corresponding author: Yuanxin Wu).*

Authors' address: Hongyan Jiang, Maoran Zhu, and Yuanxin Wu are with the Shanghai Key Laboratory of Navigation and Location-based Services, School of Electronic Information and Electrical Engineering, Shanghai Jiao Tong University, Shanghai 200240, China, E-mail: (hy_jiang@sjtu.edu.cn; zhumaoran@sjtu.edu.cn; yuanx_wu@hotmail.com).

## I. INTRODUCTION

Information pertaining to attitude, velocity, and position holds fundamental significance across various domains, such as aerospace, autonomous vehicles, computer vision, and robotics. The inertial navigation system acquires these information by integrating inertial measurements from gyroscopes and accelerometers [1-3].

The traditional inertial navigation algorithm drives from a two-speed structure, and focuses on the coning/sculling/scrolling corrections when integrating attitude/velocity/position differential equations [4], wherein the coning and sculling corrections are the dominating factors of algorithm accuracy in dynamic motions [5]. In more than half a century, a great number of researches have been devoted to attitude and velocity computation [5, 6]. Goodman and Robinson [7] proposed the simplified rotation vector differential equation of three-axis angular motions to obtain attitude by numerical integration. Savage [8] attempted to directly solve the direction cosine matrix (DCM) rate equation by the technique of two-order Picard iteration. Jordan [9] proposed the two-speed structure algorithm utilizing a combined representation of rotation vector-DCM/quaternion, wherein the high-speed computation part was founded on the simplified Goodman-Robinson rotation vector differential equation, the low-speed computation part took the rotation vector from the high-speed computation part as the input, and the DCM/quaternion representation was used for global attitude update. In the light of the earlier work by Laning [10] in 1949, Bortz [11] proposed the exact rotation vector differential equation, which, together with Jordan's two-speed structure, established the foundation of the traditional inertial navigation algorithms [12]. Miller [13] proposed that the classical coning motion should be considered as the design base for optimizing the coning correction. For the traditional velocity computation, the first-order approximation of attitude matrix has been generally adopted in the transformed specific force integration [4, 14], while a second-order approximation was highlighted by Ignagni in [2]. It is based on the exact velocity rotation compensation proposed by Savage in [14] but has not been widely exploited so far in the community.

In fact, the traditional inertial navigation algorithms possess major defects in both theory and algorithm aspects [15]. In the theory aspect, the approximate second-order integration of the rotation vector is widely used for attitude computation, and the first-order approximation of attitude matrix is commonly adopted in velocity computation. Besides, the traditional inertial navigation algorithm is usually designed under such special motion forms as the classical coning/sculling motions. Ignagni [16] noted that the influence of rotation vector approximation on a high-precision inertial navigation system could be ignored. However, the collaborative work [17] revealed the importance of the neglected high-order terms of rotation vector rate equation. Besides, Musoff [18] reported that using the classical coning motion as the algorithm design base was suboptimal under other motion forms. Notably, Savage [12] proposed a unified framework of attitude/velocity/position computation by extending the concept of rotation vector to velocity/position translation vectors. Yan [19] identified a significant problem of the error evaluation in the coning correction algorithm design, namely, increasing the number of gyroscope samples does not lead to an unlimited improvement in attitude accuracy in the traditional attitude algorithm. In the light of Taylor series expansion, Song [20] analyzed the errors of the velocity integration algorithm with the exact rotation compensation term (ViaErc) in [14] and the general velocity integration algorithm based on the simplified rotation and velocity translation vectors (ViaGen) in [12], which were both raised by Savage. In addition, Wu [21, 22] proposed the velocity/position integration formulae to thoroughly address the navigation frame rotation challenge in the traditional velocity/position computation.

The functional iteration algorithms [15, 23-25] were recently raised to directly solve the exact inertial navigation differential equations, achieving state-of-the-art ultra-high precision navigation computation. Specifically, Wu [23] introduced the functional iteration method to accurately solve the Rodrigues vector differential equation in view of its simpler form than that of the rotation vector. A faster version was further proposed in [24] by transforming the Rodrigues vector polynomial iteration into the Chebyshev polynomial coefficients iteration, where the technique of polynomial truncation was used to alleviate the computing burden without impairing the accuracy. Wu [15] further applied the functional iteration method and Chebyshev polynomial approximation to achieve the whole inertial navigation computation (iNavFIter) and diminished the noncommutativity errors to nearly machine precision. Besides, under the eight-sample scenario, Wu and Litmanovich [26] numerically compared the attitude accuracies of the Taylor series expansion and functional iteration method. Quite recently, Ignagni [5, 6] stressed the significant shortcomings of the traditional algorithms under sustained highly dynamic motions, and consequently proposed the enhanced algorithms for better accuracy.

This paper aims to analytically investigate the accuracy of the traditional algorithms and the enhanced algorithms [5, 6] against the functional iteration algorithms, centering on the accuracy improvement effect of the enhanced algorithms. The contribution of the paper is multifold. Firstly, the functional iteration approach is innovatively proposed to serve as a convenient accuracy evaluation benchmark, and the analytical accuracy evaluation of the traditional/enhanced algorithms is performed for the first time against this benchmark. The obtained error orders of algorithms align with the conclusions drawn in the existing literatures, throwing light on the moderate accuracy gains of the enhanced attitude algorithm over the traditional counterpart as well as the marginal improvement of the enhanced velocity algorithm. Notably, this benchmark enables a re-investigation of previous literature that has employed base motions having no analytical ground truth, and serves as a convenient benchmark tool to design future algorithms. Secondly, the paper reveals the discrepancy in accuracy improvements of the enhanced algorithms [5, 6] between the body-frame attitude computation scenario and the entire inertial navigation computation (attitude/velocity/position) scenario. Additionally, the extra accuracy superiority of the functional iteration algorithms under sustained lowly dynamic conditions is newly highlighted by simulation results.

The rest of the paper is structured as follows. Section II briefly summarizes the traditional and enhanced algorithms. Algorithms of interest are analyzed in the two-sample case as a demo for the sake of balance between theoretical rigor and presentation brevity. In Section III, the error orders of the traditional/enhanced algorithms, the ViaGen algorithm, and the functional iteration algorithm are analyzed and verified with the help of MATLAB Symbolic Toolbox. In Section IV, in both the two-sample and four-sample cases, all algorithms are compared against an analytical trajectory generator considering both scenarios of the body-frame attitude computation and the entire inertial navigation computation. The $1g$ cruising condition with a 50 Hz coning attitude motion is also included to mimic a vibratory environment. Conclusions are summarized in Section V.

## II. TRADITIONAL INERTIAL NAVIGATION ALGORITHMS

Suppose the computation time interval is $[0\ T]$ with any time $t$ satisfying $0 \leq t \leq T$ and $T$ denotes the time interval length. At time instants $t_k (k=1,2,\cdots)$ within the computation time interval, assuming the discrete measurements outputted by gyroscopes and accelerometers are given, the goal is to obtain the attitude, velocity, and position. The body frame is denoted by the symbol $b$, and the navigation frame is denoted by the symbol $n$. The navigation frame in this paper adopts the definition of North-Up-East [22] without loss of generality.

### A. Attitude Computation

By the chain rule of the attitude matrix, $\mathbf{C}_b^n$ at any given time can be expressed as

$$\mathbf{C}_b^n(t) = \mathbf{C}_{b(t)}^{n(t)} = \mathbf{C}_{n(0)}^{n(t)}\mathbf{C}_{b(0)}^{n(0)}\mathbf{C}_{b(t)}^{b(0)} = \mathbf{C}_{n(0)}^{n(t)}\mathbf{C}_b^n(0)\mathbf{C}_{b(t)}^{b(0)} \quad (1)$$

where $\mathbf{C}_{b(0)}^{b(t)}$ and $\mathbf{C}_{n(0)}^{n(t)}$ describe the attitude evolution of the body frame and the navigation frame from time 0 to $t$, respectively. The quaternion parameterization is chosen in this paper, namely, $\mathbf{q} = \begin{bmatrix} s & \mathbf{\eta}^T \end{bmatrix}^T$, where $s$ denotes the scalar part and $\mathbf{\eta}$ denotes the vector part. The rotation vector is represented by $\mathbf{\sigma} = \sigma \mathbf{e}$, where $\mathbf{e}$ denotes the unit vector of a fixed axis and $\sigma$ denotes the magnitude of the rotation, and then, the quaternion can be expressed by a nonzero rotation vector as follows [1]

$$\mathbf{q} = \cos\frac{\sigma}{2} + \frac{\mathbf{\sigma}}{\sigma}\sin\frac{\sigma}{2} \quad (2)$$

In view of (1), the attitude computation can be divided into two independent parts [2, 27], $\mathbf{q}_{b(0)}^{b(t)}$ and $\mathbf{q}_{n(0)}^{n(t)}$, both of which have the same function form with respect to rotation vector as (2), with the former related to $\mathbf{\sigma}_b$ and the latter related to $\mathbf{\sigma}_n$. Notably, $\mathbf{\sigma}_b$ is the rotation vector in the body frame, satisfying [11]

$$\dot{\mathbf{\sigma}}_b = \mathbf{\omega}^b + \frac{1}{2}\mathbf{\sigma}_b \times \mathbf{\omega}^b + \frac{1}{\sigma_b^2}\left[1 - \frac{\sigma_b \sin\sigma_b}{2(1-\cos\sigma_b)}\right]\mathbf{\sigma}_b \times (\mathbf{\sigma}_b \times \mathbf{\omega}^b) \quad (3)$$

where the subscript of $\mathbf{\omega}_{ib}^b$ is omitted for symbolic brevity. Besides, $\mathbf{\sigma}_n$ denotes the rotation vector in the navigation frame and satisfies the same differential equation form as in (3). However, the rotation vector in the navigation frame is usually approximated by $\mathbf{\sigma}_n \approx T\mathbf{\omega}_{in}^n$ since the navigation frame rotates rather slowly [2].

The main aim of the traditional attitude algorithm is to solve (3) for $\mathbf{\sigma}_b$. In practice, a simplified form of (3), also known as the simplified Goodman-Robinson rotation vector differential equation, is generally given as follows [7, 9, 28, 29]

$$\dot{\mathbf{\sigma}}_b = \mathbf{\omega}^b + \frac{1}{2}\mathbf{\alpha} \times \mathbf{\omega}^b \quad (4)$$

where the angular increment is defined as $\mathbf{\alpha} = \int_0^t \mathbf{\omega}^b \, dt$. Suppose the angular velocity is expressed as a polynomial of time as $\mathbf{\omega}^b = \mathbf{a}_\omega + \mathbf{b}_\omega t$, where $\mathbf{a}_\omega$ and $\mathbf{b}_\omega$ are constant vector coefficients. Note that the above linear form is used just for a balance between theoretical rigor and presentation brevity. Then the approximate rotation vector can be obtained by integrating (4) on the computation time interval, say

$$\tilde{\mathbf{\sigma}}_b = \mathbf{a}_\omega t + \mathbf{b}_\omega \frac{t^2}{2} + \mathbf{a}_\omega \times \mathbf{b}_\omega \frac{t^3}{12} \quad (5)$$

The above polynomial coefficients can be readily obtained using the angular increments measured by gyroscopes, see e.g. [21], and thus $\tilde{\mathbf{\sigma}}_b$ can be readily solved.

Ignagni [2, 6] analyzed the theoretical errors implicit in the simplified Goodman-Robinson rotation vector differential equation and proposed an enhanced attitude algorithm. Specifically, the approximate rotation vector $\tilde{\mathbf{\sigma}}_b$ is substituted into the right side of (3), and the scalar coefficient of the third term is approximated as 1/12, say

$$\dot{\mathbf{\sigma}}_b = \mathbf{\omega}^b + \frac{1}{2}\tilde{\mathbf{\sigma}}_b \times \mathbf{\omega}^b + \frac{1}{12}\tilde{\mathbf{\sigma}}_b \times (\tilde{\mathbf{\sigma}}_b \times \mathbf{\omega}^b) \quad (6)$$

Similarly, the refined rotation vector $\mathbf{\sigma}_b$ is obtained as $\mathbf{\sigma}_b \triangleq \tilde{\mathbf{\sigma}}_b + \delta\mathbf{\sigma}_b$, where the compensated rotation vector part, $\delta\mathbf{\sigma}_b$, is defined as follows

$$\begin{aligned}
\delta\mathbf{\sigma}_b &= \left[(\mathbf{a}_\omega \times \mathbf{b}_\omega) \times \mathbf{b}_\omega + \frac{1}{3}\mathbf{a}_\omega \times ((\mathbf{a}_\omega \times \mathbf{b}_\omega) \times \mathbf{a}_\omega)\right]\frac{t^5}{240} \\
&+ \left[\mathbf{a}_\omega \times ((\mathbf{a}_\omega \times \mathbf{b}_\omega) \times \mathbf{b}_\omega) + \frac{1}{2}\mathbf{b}_\omega \times ((\mathbf{a}_\omega \times \mathbf{b}_\omega) \times \mathbf{a}_\omega)\right]\frac{t^6}{864} \\
&+ \left[\mathbf{b}_\omega \times ((\mathbf{a}_\omega \times \mathbf{b}_\omega) \times \mathbf{b}_\omega) + \frac{1}{6}(\mathbf{a}_\omega \times \mathbf{b}_\omega) \times ((\mathbf{a}_\omega \times \mathbf{b}_\omega) \times \mathbf{a}_\omega)\right]\frac{t^7}{2016} \\
&+ (\mathbf{a}_\omega \times \mathbf{b}_\omega) \times ((\mathbf{a}_\omega \times \mathbf{b}_\omega) \times \mathbf{b}_\omega)\frac{t^8}{13824}
\end{aligned} \quad (7)$$

It should be highlighted that the algorithmic attitude error of traditional attitude algorithms, as defined in [6] as the errors in integrating the simplified rotation-vector equation, depends on the specific optimized coning-compensation algorithm. For example, the algorithmic error vanishes in the two-sample case (the coefficient $C$ is zero in Table 2 of [30]) but does not otherwise.

**B. Velocity Computation**

In terms of traditional velocity computation, the velocity update equation is shown as follows [2, 27]

$$\begin{aligned}
\mathbf{v}^n &\approx \mathbf{v}^n(0) + \mathbf{C}_{b(0)}^{n(0)}\int_0^t \mathbf{C}_{b(t)}^{b(0)}\mathbf{f}^b \, dt \\
&- \int_0^t (2\mathbf{\omega}_{ie}^n + \mathbf{\omega}_{en}^n) \times \mathbf{v}^n \, dt + \int_0^t \mathbf{g}^n \, dt + \Delta\mathbf{v}_f^c \\
&= \mathbf{v}^n(0) + \mathbf{C}_{b(0)}^{n(0)}\int_0^t \left[\mathbf{I} + \frac{\sin\sigma_b}{\sigma_b}(\mathbf{\sigma}_b \times) + \frac{1-\cos\sigma_b}{\sigma_b^2}(\mathbf{\sigma}_b \times)^2\right]\mathbf{f}^b \, dt \\
&- \int_0^t (2\mathbf{\omega}_{ie}^n + \mathbf{\omega}_{en}^n) \times \mathbf{v}^n \, dt + \int_0^t \mathbf{g}^n \, dt + \Delta\mathbf{v}_f^c
\end{aligned} \quad (8)$$

where the integration of the transformed specific force is denoted as $\mathbf{C}_{b(0)}^{n(0)}\int_0^t \mathbf{C}_{b(t)}^{b(0)}\mathbf{f}^b \, dt \triangleq \mathbf{u}$, and $\Delta\mathbf{v}_f^c$ is the compensation term taking account of the navigation-frame rotation effect over the computational interval, defined by [2, 5, 14]

$$\Delta \mathbf{v}_f^c = -\frac{T}{2}\left(\boldsymbol{\omega}_{ie}^n + \boldsymbol{\omega}_{en}^n\right) \times \left(\mathbf{C}_{b(0)}^{n(0)}\boldsymbol{\upsilon}\right) \quad (9)$$

where the velocity increment is defined as $\boldsymbol{\upsilon} = \int_0^t \mathbf{f}^b \, dt$. Generally, the gravity vector and Coriolis acceleration change slowly during the usually short computing time interval, so the sum of the penultimate and antepenultimate terms of (8) can be approximated as [2, 27]

$$\Delta \mathbf{v}_g \triangleq -\int_0^t \left(2\boldsymbol{\omega}_{ie}^n + \boldsymbol{\omega}_{en}^n\right) \times \mathbf{v}^n \, dt + \int_0^t \mathbf{g}^n \, dt \quad (10)$$
$$\approx \left[\mathbf{g}^n - \left(2\boldsymbol{\omega}_{ie}^n + \boldsymbol{\omega}_{en}^n\right) \times \mathbf{v}^n\right] t$$

Typically, the simplified velocity update equation, up to the first-order angular increment, is given as [2, 4, 12, 14]

$$\mathbf{v}^n = \mathbf{v}^n(0) + \mathbf{C}_{b(0)}^{n(0)} \int_0^t \left(\mathbf{I} + (\boldsymbol{\alpha} \times)\right) \mathbf{f}^b dt + \Delta \mathbf{v}_g + \Delta \mathbf{v}_f^c \quad (11)$$

where $\int_0^t \boldsymbol{\alpha} \times \mathbf{f}^b \, dt$ is the commonly-used first-order velocity correction [2]. For the same brevity consideration, assume that the specific force is also represented as a linear polynomial of time, $\mathbf{f}^b = \mathbf{a}_f + \mathbf{b}_f t$, where $\mathbf{a}_f$ and $\mathbf{b}_f$ are constant vector coefficients. The integral term of (11) can be rewritten as [21]

$$\int_0^t \left(\mathbf{I} + (\boldsymbol{\alpha} \times)\right) \mathbf{f}^b \, dt$$
$$= \boldsymbol{\upsilon} + \frac{1}{2}\left(\mathbf{a}_\omega \times \mathbf{a}_f\right)t^2 + \frac{1}{3}\left(\frac{1}{2}\mathbf{b}_\omega \times \mathbf{a}_f + \mathbf{a}_\omega \times \mathbf{b}_f\right)t^3 \quad (12)$$
$$+ \frac{1}{8}\left(\mathbf{b}_\omega \times \mathbf{b}_f\right)t^4$$

On the other hand, the simplified velocity update equation, up to the second-order angular increment, is given as [2, 5]

$$\mathbf{v}^n = \mathbf{v}^n(0) + \mathbf{C}_{b(0)}^{n(0)} \int_0^t \left[\mathbf{I} + (\boldsymbol{\alpha} \times) + \frac{1}{2}(\boldsymbol{\alpha} \times)^2\right] \mathbf{f}^b \, dt \quad (13)$$
$$+ \Delta \mathbf{v}_g + \Delta \mathbf{v}_f^c$$

The term $\int_0^t \frac{1}{2}(\boldsymbol{\alpha} \times)^2 \mathbf{f}^b \, dt$ is called the second-order velocity correction [2], where $\boldsymbol{\omega}^b$ and $\mathbf{f}^b$ are further approximately considered as constants. Thus the simplified second-order velocity correction is calculated as [2, 5]

$$\int_0^t \frac{1}{2}(\boldsymbol{\alpha} \times)^2 \mathbf{f}^b \, dt \approx \frac{1}{6}\boldsymbol{\alpha} \times (\boldsymbol{\alpha} \times \boldsymbol{\upsilon}) \quad (14)$$

Therefore, the second-order velocity update equation is further approximated as follows [2, 5]

$$\mathbf{v}^n = \mathbf{v}^n(0) + \mathbf{C}_{b(0)}^{n(0)}\left[\int_0^t \left(\mathbf{I} + (\boldsymbol{\alpha} \times)\right)\mathbf{f}^b \, dt + \frac{1}{6}\boldsymbol{\alpha} \times (\boldsymbol{\alpha} \times \boldsymbol{\upsilon})\right] \quad (15)$$
$$+ \Delta \mathbf{v}_g + \Delta \mathbf{v}_f^c$$

In the case of two-sample algorithm, the term in the square bracket of (15) can be represented by constant vector coefficients $\mathbf{a}_\omega$, $\mathbf{b}_\omega$, $\mathbf{a}_f$ and $\mathbf{b}_f$. The specific expression is not explicitly given here to avoid verbosity.

Hereafter, the algorithms based on the velocity update equation (11) and (15) are called the traditional first-order and second-order algorithms, respectively. The latter shares the same spirit as the ViaErc algorithm [14, 20]. It can be readily shown that the ViaErc algorithm is equivalent to the traditional second-order algorithm in (15).

It is obvious that both the first-order and second-order velocity corrections are undermined with potentially significant approximation errors, which were recently reiterated and compensated by the enhanced velocity algorithms in [2, 5]. The theoretical error in the first-order velocity correction is [2, 5]

$$\delta \mathbf{v}_1 = \int_0^t \left[\tilde{\boldsymbol{\sigma}}_b - \boldsymbol{\alpha}\right] \times \mathbf{f}^b \, dt \quad (16)$$

and the theoretical error in the second-order velocity correction is [2, 5]

$$\delta \mathbf{v}_2 = \frac{1}{2}\int_0^t \tilde{\boldsymbol{\sigma}}_b \times \left[\tilde{\boldsymbol{\sigma}}_b \times \mathbf{f}^b\right] dt - \boldsymbol{\alpha} \times (\boldsymbol{\alpha} \times \boldsymbol{\upsilon})/6 \quad (17)$$

For the linear motion form, the theoretical errors implicit in the first-order and the second-order velocity corrections can be specifically derived as follows, respectively,

$$\delta \mathbf{v}_1 = (\mathbf{a}_\omega \times \mathbf{b}_\omega) \times \mathbf{a}_f \frac{t^4}{48} + (\mathbf{a}_\omega \times \mathbf{b}_\omega) \times \mathbf{b}_f \frac{t^5}{60} \quad (18)$$

$$\delta \mathbf{v}_2$$
$$= \left[2\mathbf{a}_\omega \times (\mathbf{a}_\omega \times \mathbf{b}_f) - \mathbf{a}_\omega \times (\mathbf{b}_\omega \times \mathbf{a}_f) - \mathbf{b}_\omega \times (\mathbf{a}_\omega \times \mathbf{a}_f)\right]\frac{t^4}{48}$$
$$+ \left\{\mathbf{a}_\omega \times (\mathbf{b}_\omega \times \mathbf{b}_f) + \mathbf{a}_\omega \times \left[(\mathbf{a}_\omega \times \mathbf{b}_\omega) \times \mathbf{a}_f\right] + \mathbf{b}_\omega \times (\mathbf{a}_\omega \times \mathbf{b}_f)\right.$$
$$\left. -2\mathbf{b}_\omega \times (\mathbf{b}_\omega \times \mathbf{a}_f) + (\mathbf{a}_\omega \times \mathbf{b}_\omega) \times (\mathbf{a}_\omega \times \mathbf{a}_f)\right\}\frac{t^5}{120}$$
$$+ \left\{\mathbf{a}_\omega \times \left[(\mathbf{a}_\omega \times \mathbf{b}_\omega) \times \mathbf{b}_f\right] + \frac{1}{2}\mathbf{b}_\omega \times \left[(\mathbf{a}_\omega \times \mathbf{b}_\omega) \times \mathbf{a}_f\right]\right.$$
$$\left. +(\mathbf{a}_\omega \times \mathbf{b}_\omega) \times (\mathbf{a}_\omega \times \mathbf{b}_f) + \frac{1}{2}(\mathbf{a}_\omega \times \mathbf{b}_\omega) \times (\mathbf{b}_\omega \times \mathbf{a}_f)\right\}\frac{t^6}{144}$$
$$+ \left\{\mathbf{b}_\omega \times \left[(\mathbf{a}_\omega \times \mathbf{b}_\omega) \times \mathbf{b}_f\right] + (\mathbf{a}_\omega \times \mathbf{b}_\omega) \times (\mathbf{b}_\omega \times \mathbf{b}_f)\right.$$
$$\left. +\frac{1}{6}(\mathbf{a}_\omega \times \mathbf{b}_\omega) \times \left[(\mathbf{a}_\omega \times \mathbf{b}_\omega) \times \mathbf{a}_f\right]\right\}\frac{t^7}{336}$$
$$+ (\mathbf{a}_\omega \times \mathbf{b}_\omega) \times \left[(\mathbf{a}_\omega \times \mathbf{b}_\omega) \times \mathbf{b}_f\right]\frac{t^8}{2304}$$
(19)

Hence, the velocity update equations of the traditional first-order and second-order algorithms are respectively enhanced as follows [5, 6]

$$\mathbf{v}^n = \mathbf{v}^n(0) + \mathbf{C}_{b(0)}^{n(0)}\left[\int_0^t \left[\mathbf{I} + (\boldsymbol{\alpha} \times)\right]\mathbf{f}^b dt + \delta \mathbf{v}_1\right] + \Delta \mathbf{v}_g + \Delta \mathbf{v}_f^c$$
(20)

$$\mathbf{v}^n = \mathbf{v}^n(0) + \mathbf{C}_{b(0)}^{n(0)} \left[ \int_0^t \left[ \mathbf{I} + (\boldsymbol{\alpha} \times) \right] \mathbf{f}^b \, dt + \frac{1}{6} \boldsymbol{\alpha} \times (\boldsymbol{\alpha} \times \boldsymbol{\upsilon}) \right.$$
$$\left. + \delta \mathbf{v}_1 + \delta \mathbf{v}_2 \right] + \Delta \mathbf{v}_g + \Delta \mathbf{v}_f^c$$
(21)

Hereafter, (20) and (21) are called the enhanced first-order and second-order algorithms, respectively. By analogy, the algorithmic errors of the traditional velocity algorithm were also defined in [5] that vanishes as well for the two-sample case (the coefficients **d** and **g** are zeros in Table 3 of [31]). Due to the page limit, the subsequent sections will focus on the traditional/enhanced second-order algorithms.

Furthermore, a brief review of the ViaGen algorithm is also given below for a comprehensive accuracy comparison. Specifically, the integration of the transformed specific force therein is [14, 20]

$$\mathbf{u}_{Gen}$$
$$= \mathbf{C}_{b(0)}^{n(0)} \left[ \boldsymbol{\eta} + \frac{1-\cos\tilde{\sigma}_b}{\tilde{\sigma}_b^2} \tilde{\boldsymbol{\sigma}}_b \times \boldsymbol{\eta} + \frac{1}{\tilde{\sigma}_b^2}\left(1 - \frac{\sin\tilde{\sigma}_b}{\tilde{\sigma}_b}\right) \tilde{\boldsymbol{\sigma}}_b \times (\tilde{\boldsymbol{\sigma}}_b \times \boldsymbol{\eta}) \right]$$
$$\approx \mathbf{C}_{b(0)}^{n(0)} \left[ \boldsymbol{\eta} + \frac{1}{2} \tilde{\boldsymbol{\sigma}}_b \times \boldsymbol{\eta} + \frac{1}{6} \tilde{\boldsymbol{\sigma}}_b \times (\tilde{\boldsymbol{\sigma}}_b \times \boldsymbol{\eta}) \right]$$
(22)

where the velocity translation vector $\boldsymbol{\eta}$ is approximately given as [14, 20]

$$\boldsymbol{\eta} \approx \boldsymbol{\upsilon} + \frac{1}{2} \int_0^t (\boldsymbol{\alpha} \times \mathbf{f}^b - \boldsymbol{\omega} \times \boldsymbol{\upsilon}) \, dt \quad (23)$$

As for the position update, an algorithmic compensation scheme was given in [5], wherein the Simpson formula was used instead of the trapezoidal formula to compute the integral of the velocity. Because this scheme is tightly related to the specific definition of update interval, and does not focus on the realm of compensating for implicit errors, we do not take a closer look into the compensation scheme of position update in this paper.

## III. ALGORITHM ACCURACY ANALYSIS

In this section, we accomplish the algorithm accuracy analysis by the aid of MATLAB Symbolic Toolbox, in contrast to the traditional manual mathematical derivation. By so doing, it helps us to spare the tedious mathematical details and concentrate on the main idea. Note that no compensation for the theoretical error implicit in the traditional position update equation was discussed in [5], so this paper focuses on the attitude and velocity computation accuracy.

The functional iteration approach is capable of provably solving the exact attitude and velocity update equations [15, 23, 25] and consequently obtaining the exact analytical solution of the true rotation vector and the true velocity. In this regard, the error orders of the traditional/enhanced algorithms and the ViaGen algorithm can be analyzed against the functional iteration approach with the help of MATLAB Symbolic Toolbox, aiming to study the potential accuracy advantages of the enhanced algorithms and the ViaGen algorithm over the traditional ones. It is important to mention that, regarding the error orders of the attitude algorithms, here we only focus on the accuracy of inertial attitude computation, specifically the solution of (3). Additionally, in terms of the error orders of the velocity algorithms, particular attention is paid to the integration of the transformed specific force (denoted as $\mathbf{u}$ below (8)). Notably, when the functional iteration approach is used to solve (3), the trigonometric function coefficient in the third term is expanded to the eighth-order Taylor series [26].

In the MATLAB symbolic development, the coefficients $\mathbf{a}_\omega$, $\mathbf{b}_\omega$, $\mathbf{a}_f$ and $\mathbf{b}_f$ in Section II are all generated as random vectors to make sure the derived conclusion has a general sense. However, in the sequel we just exemplify a specific set of vector coefficients for concise demonstration. It should be highlighted that the following observations are consistent throughout all of our random vector coefficients. Specifically, we set $\mathbf{a}_\omega = \begin{bmatrix} 4 & 2 & 3 \end{bmatrix}^T$, $\mathbf{b}_\omega = \begin{bmatrix} 5 & 8 & 10 \end{bmatrix}^T$, $\mathbf{a}_f = \begin{bmatrix} 4 & 5 & 6 \end{bmatrix}^T$, and $\mathbf{b}_f = \begin{bmatrix} 9 & 8 & 7 \end{bmatrix}^T$. The resultant polynomial coefficients of essential attitude and velocity computation are listed in Tables I and II, respectively. The attitude coefficients convergence process of the functional iteration approach across eight iterations is also shown in Table I.

With the result of the functional iteration approach as the reference, Table I indicates that the error order of the traditional attitude algorithm is $O(t^5)$, consistent with the conclusion in [6], and that of the enhanced attitude algorithm is $O(t^6)$, indicating the moderate accuracy gains of the enhanced attitude algorithm over the traditional counterpart. Moreover, Table II shows that the error orders of the traditional/enhanced second-order velocity algorithms are both $O(t^4)$, implying the marginal accuracy improvement of enhanced velocity algorithm over the traditional one. In addition, when the eight-order Taylor series of trigonometric coefficients in (22) is used, Table II indicates that the error order of the ViaGen algorithm is $O(t^5)$, agreeing with the analysis in [20]. However, when utilizing the first-order simplification of trigonometric coefficients, the accuracy of the ViaGen algorithm decreases by one order of time. Moreover, the ViaGen algorithm with an eight-order series of trigonometric coefficients demonstrates one order of time higher accuracy than both the traditional/enhanced second-order algorithms, highlighting the positive accuracy impact

**Table I**
**Polynomial Coefficients of Inertial Attitude Computation in a Test Example**

|  | $t$ | $t^2$ | $t^3$ | $t^4$ | $t^5$ | $t^6$ | $t^7$ | $t^8$ |
|---|---|---|---|---|---|---|---|---|
| Typical | 4 | 5/2 | -1/3 | 0 | 0 | 0 | 0 | 0 |
| Enhanced | 4 | 5/2 | -1/3 | 0 | -697/360 | -11/24 | -251/336 | -625/1536 |
| FIterTrue ($l$=1) | 4 | 5/2 | 0 | 0 | 0 | 0 | 0 | 0 |
| FIterTrue ($l$=2) | 4 | 5/2 | -1/3 | 119/96 | 71/40 | 3451/8640 | 1559/1120 | 264763/138240 |
| FIterTrue ($l$=3) | 4 | 5/2 | -1/3 | 0 | -1481/720 | 153/1280 | 88181/40320 | 51048661/11612160 |
| FIterTrue ($l$=4) | 4 | 5/2 | -1/3 | 0 | -697/360 | -1535/2304 | -213839/60480 | -56149561/7741440 |
| FIterTrue ($l$=5) | 4 | 5/2 | -1/3 | 0 | -697/360 | -11/30 | -39065/24192 | -13712053/3870720 |
| FIterTrue ($l$=6) | 4 | 5/2 | -1/3 | 0 | -697/360 | -11/30 | -3533/2160 | -8644241/2580480 |
| FIterTrue ($l$=7,8) | 4 | 5/2 | -1/3 | 0 | -697/360 | -11/30 | -3533/2160 | -13663/4032 |

Note: '*Typical*' denotes the traditional attitude algorithm, '*Enhanced*' denotes the enhanced attitude algorithm, '*FIterTrue*' denotes the functional iteration attitude algorithm that is used as the reference, and $l$ denotes the functional iteration times.

**Table II**
**Polynomial Coefficients of Essential Velocity Computation of u in a Test Example**

|  | $t$ | $t^2$ | $t^3$ | $t^4$ | $t^5$ | $t^6$ | $t^7$ | $t^8$ |
|---|---|---|---|---|---|---|---|---|
| Typical | 4 | 3 | 19/3 | 167/12 | -263/24 | -403/24 | 0 | 0 |
| Enhanced | 4 | 3 | 19/3 | 149/24 | -253/15 | -6601/288 | -935/168 | -9797/2304 |
| ViaGen-1 | 4 | 3 | 19/3 | 149/24 | -1685/144 | -1517/96 | 241/216 | -785/216 |
| ViaGen-8 | 4 | 3 | 19/3 | 59/6 | -1541/144 | -41113/1440 | -129137/60480 | 0 |
| FIterTrue | 4 | 3 | 19/3 | 59/6 | -212/15 | -4987/180 | 7681/630 | 24079/360 |

Note: '*Typical*' denotes the traditional second-order velocity algorithm, '*Enhanced*' denotes the enhanced second-order velocity algorithm, '*ViaGen-1*' denotes the ViaGen velocity algorithm with first-order series of trigonometric coefficients, '*ViaGen-8*' denotes the ViaGen velocity algorithm with eight-order series of trigonometric coefficients, and '*FIterTrue*' denotes the functional iteration velocity algorithm that is used as the reference.

of the new concept of velocity translation vector.

Note that the above-obtained conclusions are not limited to the first-order polynomial functions of angular velocity/specific force with respect to time $t$, and they remain valid for high-order polynomials according to our extra tests, e.g., the third-order polynomials. However, due to page constraints, further details on high-order polynomials are not provided here.

## IV. SIMULATION RESULTS

In this section, we simulate global fight test data sets following our previous work [15], assessing numerical outputs of all considered algorithms against the analytical trajectory generator. For a comprehensive accuracy comparison and evaluation with [5, 6], in both the two-sample and four-sample cases, we concurrently assess the performance of traditional algorithms, enhanced algorithms, and functional iteration algorithms, considering both scenarios of the body-frame attitude computation and the entire inertial navigation computation. Notably, both implicit and algorithmic errors of the traditional algorithms are taken into account [5, 6]. The four-sample coning correction of *Algorithm 7* in [30] and the four-sample sculling correction of *Algorithm 5* in [31] are employed. As for the functional iteration algorithms, the maximum degrees of Chebyshev polynomials used to fit the angular velocity, specific force and gravity vector as well as the truncation degrees of attitude, velocity, and position are set to be the same as [15]. The convergence criterion of iteration is set to $10^{-16}$, and the maximum number of attitude iteration and velocity/position iteration are set as $N+1$, where N denotes the number of samples.

For the body-frame attitude computation scenario, simulations are performed under coning motion, which has analytical expressions in angular velocity and the associated attitude quaternion [13, 25], and it is universally employed as the most demanding test input in evaluating

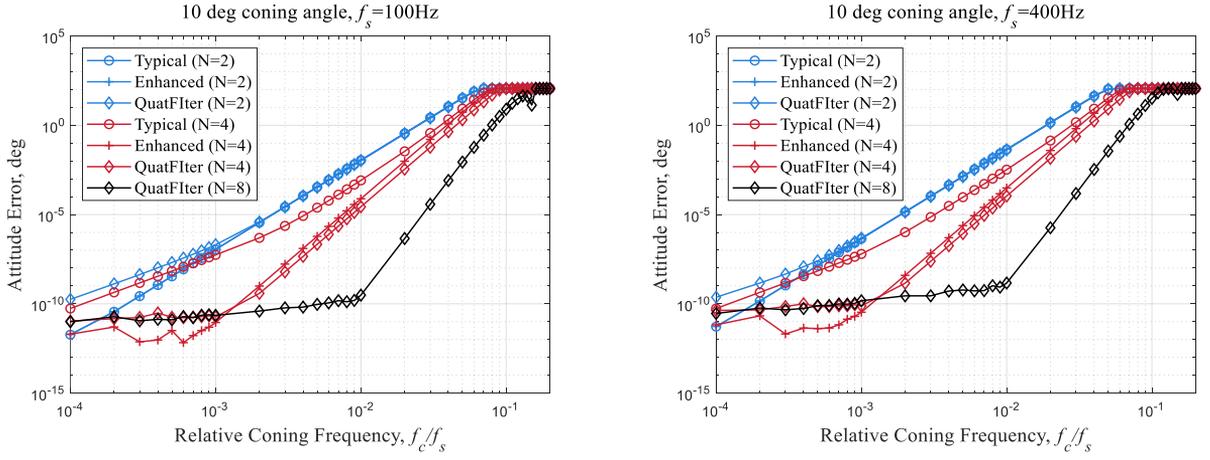

Fig. 1. Principal angle errors as a function of relative frequency for 100 Hz and 400 Hz sampling frequencies ('*Typical*' denotes the traditional attitude algorithm, '*Enhanced*' denotes the enhanced attitude algorithm, '*QuatFIter*' denotes the quaternion functional iteration algorithm; N denotes the number of samples).

attitude accuracy. The angular velocity of the coning motion is given by [25]

$$\boldsymbol{\omega} = \Omega[-2\sin^2(\zeta/2) \ -\sin(\zeta)\sin(\Omega t) \ \sin(\zeta)\cos(\Omega t)]^T \quad (24)$$

where $\zeta$ is the coning angle (unit: deg), $\Omega = 2\pi f_c$ is the angular frequency of the coning motion (unit: rad/s), and $f_c$ denotes the coning frequency (unit: Hz). The coning angle is set to $\zeta = 10°$, and both 100 Hz and 400 Hz sampling frequencies are considered. The principal angle metric is employed to evaluate the attitude computation error [25]. Notably, since only attitude computation is concerning, the functional iteration algorithm here specifically refers to the QuatFIter algorithm [25]. In order to evaluate the algorithms under a range of different dynamic conditions, for sampling frequencies of 100 Hz and 400 Hz, the coning frequencies are set to vary within the ranges of 0.01-20 Hz and 0.04-80 Hz, respectively. The maximum principal angle errors are analyzed over a data duration of 4000 seconds.

Fig. 1 presents the error results as a function of relative frequency in the body-frame attitude computation scenario. Regardless of whether the sampling frequency is set to 100 Hz or 400 Hz, it is evident that all considered algorithms exhibit identical accuracy rankings. In specific, for the case of two samples, it is observed that both traditional and enhanced two-sample algorithms have the same accuracy across the considered entire dynamic range. It is crucial to note that the limited number of two samples introduce significant angular velocity fitting errors, thus impairing the accuracy of the functional iteration algorithm (N=2) significantly in the lower range of relative frequency. Additionally, for the case of four samples, the enhanced four-sample algorithm demonstrates noticeable accuracy improvement over the traditional algorithms, particularly in the lower range of relative frequency. Moreover, within the relative frequency range of $10^{-3}$ to $10^{-1}$, the accuracy of the enhanced four-sample algorithm is comparable with that of the functional iteration four-sample algorithm, with the latter marginally better. It is worth noting that within the relative frequency range of $10^{-4}$ to $10^{-3}$, the enhanced four-sample algorithm performs marginally better than the functional iteration algorithms, in which the interpolation errors of angular velocity from angular increments play a dominating negative role in functional iteration while the enhanced four-sample algorithm directly uses the exact angular increments. However, within this dynamic range, these two algorithms already reach extremely low level errors, literally negligible for practical applications. With the increase of the number of samples, the accuracy superiority of the functional iteration algorithms becomes more significant, as illustrated by the curve of QuatFIter (N=8) in the figures.

For the entire inertial navigation computation scenario, the fight test simulation in line with our previous works [15, 22, 26] is used for algorithms evaluation, where data is generated in the navigation frame and the analytical truth is favorably available for accuracy evaluation. Specifically, the vehicle is initially located at zero longitude/latitude/height, and moves eastward with an initial speed $v_0 = 500 \text{ m/s}$ and velocity rate $\dot{\mathbf{v}}^n = \begin{bmatrix} 0 & 0 & a\sin(wt) \end{bmatrix}^T$, where $a$ and $w$ represents the magnitude and angular frequency of the velocity rate, respectively ($a = 10 \text{ m/s}^2$, $w = 0.02 \text{ rad/s}$). Additionally, the body attitude undergoes a classical coning motion, which is described by attitude quaternion as [1, 13]

$$\mathbf{q}_n^b = \cos(\zeta/2) + \sin(\zeta/2)\begin{bmatrix} 0 & \cos(\Omega t) & \sin(\Omega t) \end{bmatrix}^T \quad (25)$$

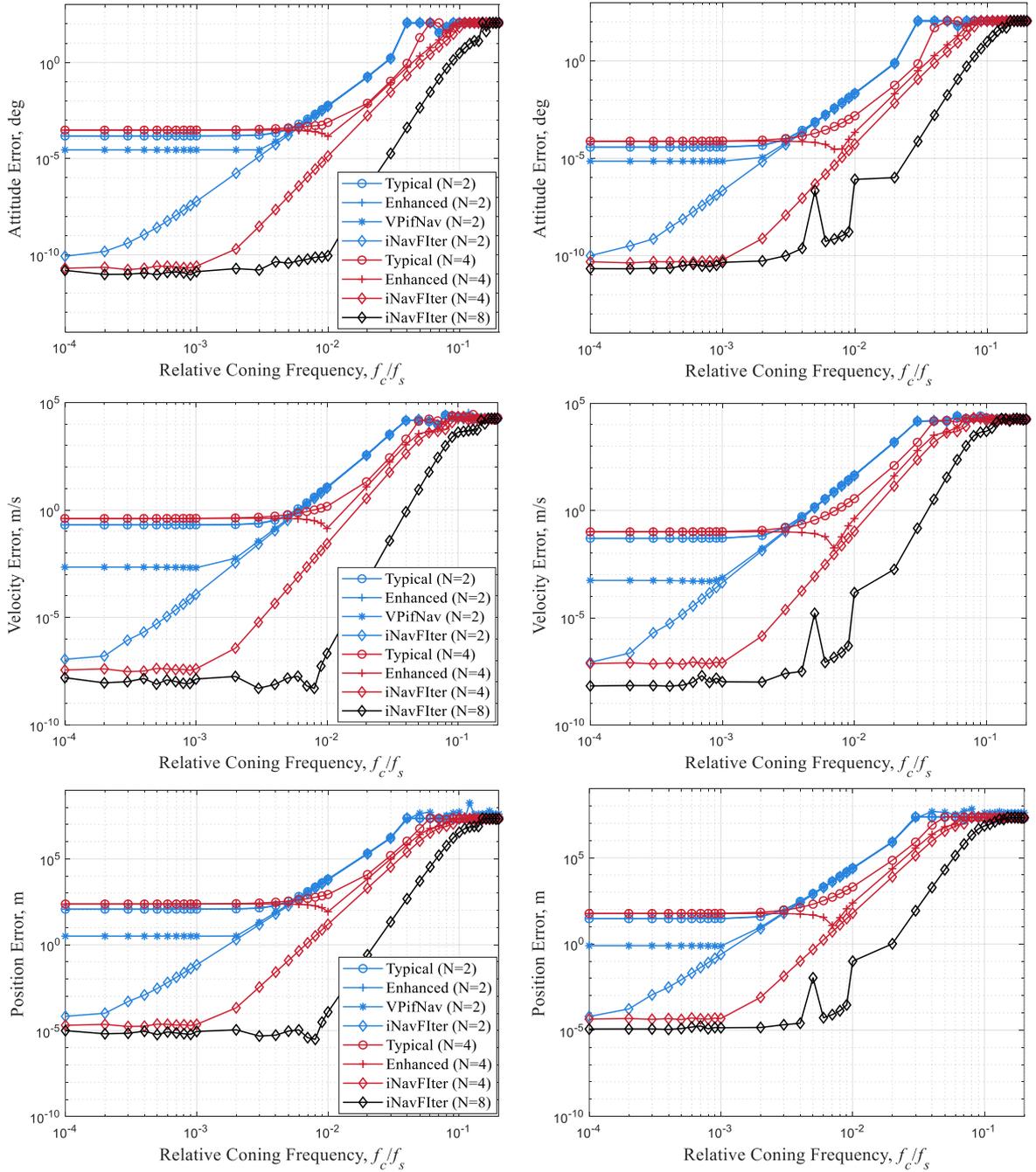

Fig. 2. Principal angle error, the magnitude of velocity error and position error as a function of relative frequency (left column for 100 Hz sampling frequency, right column for 400 Hz sampling frequency, 10 degree coning angle; '*Typical*' denotes the traditional navigation algorithm, '*Enhanced*' denotes the enhanced navigation algorithm, '*iNavFIter*' denotes the functional iteration navigation algorithm, '*VPifNav*' denotes the navigation algorithm especially considering reference frame rotation; N denotes the number of samples)

where the coning angle and sampling frequencies are set as in the body-frame attitude computation scenario. Note that the traditional/enhanced algorithms here are based on the second-order velocity update equations and take into account the compensation for the navigation-frame rotation effect as well. The velocity/position integration formula with navigation reference frame rotation considered (named the VPifNav algorithm [22]) is also compared to highlight the significance of navigation reference frame rotation effect.

The accuracy performance of all algorithms is investigated within the same dynamic range as in the body-frame attitude computation scenario, and an analysis is

**Table III**

**Summary of Maximum West-East Position Errors at Representative Coning Frequencies**.

|  | $f_c = 0.037$ Hz | $f_c = 0.185$ Hz | $f_c = 1$ Hz |
|---|---|---|---|
| Typical (N=2) | 16.83 m | 16.95 m | 965.18 m |
| Enhanced (N=2) | 16.83 m | 16.80 m | 980.98 m |
| VPifNav (N=2) | 3.17 m | 2.93 m | 970.28 m |
| iNavFIter (N=2) | $7.34 \times 10^{-5}$ m | 0.20 m | 929.31 m |
| Typical (N=4) | 33.72 m | 34.85 m | 137.88 m |
| Enhanced (N=4) | 33.71 m | 34.11 m | 24.63 m |
| iNavFIter (N=4) | $3.37 \times 10^{-6}$ m | $1.35 \times 10^{-5}$ m | 2.40 m |
| iNavFIter (N=8) | $4.27 \times 10^{-6}$ m | $4.36 \times 10^{-6}$ m | $2.05 \times 10^{-5}$ m |

conducted on the maximum principal angle error, maximum magnitudes of velocity and position errors over a data duration of 4000 seconds. Fig. 2 plots the error results as a function of relative frequency at sampling frequencies of both 100 Hz and 400 Hz in the entire inertial navigation computation scenario, and Table III lists the specific maximum west-east position errors at three representative coning frequencies under 100 Hz sampling frequency for demonstration.

It can be seen that all algorithms still perform by the same accuracy rankings for both sampling frequencies. Below, the specifics are elucidated using the case of 100 Hz sampling frequency, with similar conclusions drawn for the 400 Hz case. Firstly, the focus is on the accuracy comparison between the enhanced algorithms and the traditional algorithms. In the two-sample case, the enhanced two-sample algorithm achieves nearly identical accuracy to the traditional counterpart. In the four-sample case, the enhanced four-sample algorithm also achieves the same level of accuracy as the traditional counterpart when the relative frequency is below around $5 \times 10^{-3}$, but beyond this threshold, the enhanced four-sample algorithm shows a marginal improvement, particularly within the relative frequency range of $5 \times 10^{-3}$ to $2 \times 10^{-2}$. Moreover, comparing Figs. 1-2, it is observed that the attitude improvement of the enhanced four-sample algorithm over the traditional one significantly shrinks in the entire inertial navigation computation scenario. The reason lies in the insufficiency of the traditional and enhanced velocity/position computation, especially in the integration of the transformed specific force and the computational reference frame rotation compensation. Besides, in the entire inertial navigation computation scenario, attitude computation involves computing not only the rotation vector of the body frame but also that of the navigation frame, the latter of which introduces substantial approximation errors as well and further limits or overshadows the overall accuracy improvement of enhanced algorithms.

In addition, concerning the accuracy performance of the functional iteration algorithms, at low relative frequencies, the functional iteration two-sample algorithm exhibits a substantial accuracy superiority over both the traditional/enhanced two-sample and four-sample algorithms. This superiority becomes more pronounced with an increasing number of samples, as evidenced by the curves of iNavFIter (N=4) and iNavFIter (N=8) in Fig. 2. However, at high coning frequencies, the accuracies of the functional iteration two-sample algorithm and the traditional/enhanced two-sample algorithms are comparable, and so are the corresponding four-sample algorithms. The underlying reason is that whether two samples or four samples are used to fit the measurements of gyroscopes and accelerometers, they would encounter significant angular velocity/specific force fitting errors at high coning frequencies, which dominates the inertial navigation computation errors in any algorithm (including the functional iteration). Therefore, for a high coning frequency, more samples are preferred to reduce the fitting errors as much as possible, as shown by the black-diamond curves in Fig. 2 for iNavFIter (N=8). Nevertheless, the number of samples cannot be increased without limit, as the interpolation quality on the equispaced sampling points is affected by the Runge phenomenon [32].

Furthermore, it may be surprising at first glance that the traditional/enhanced algorithms have relatively high errors in lowly dynamic conditions. The primary reason stems from the insufficient refinement in the treatment of components related to the navigation frame in the attitude and velocity update equations of traditional/enhanced algorithms. In particular, the insufficient compensation of the navigation reference frame rotation effect in traditional/enhanced velocity equation results in considerable error, as evidenced by the curves of the VPifNav algorithm in Fig. 2. However, along with the increase of coning frequency, the dynamics of the body frame is strengthened and the errors caused by the above factors are no longer dominant, which explains the curves coincidence of the traditional/enhanced two-sample algorithms and the VPifNav algorithm at high coning frequencies. Surprisingly, the traditional/enhanced two-sample algorithms slightly outperform the four-sample counterparts at lower frequencies. We speculate that this phenomenon is caused by a halved update frequency compared to the two-sample case that affects the navigation frame related attitude/velocity computation.

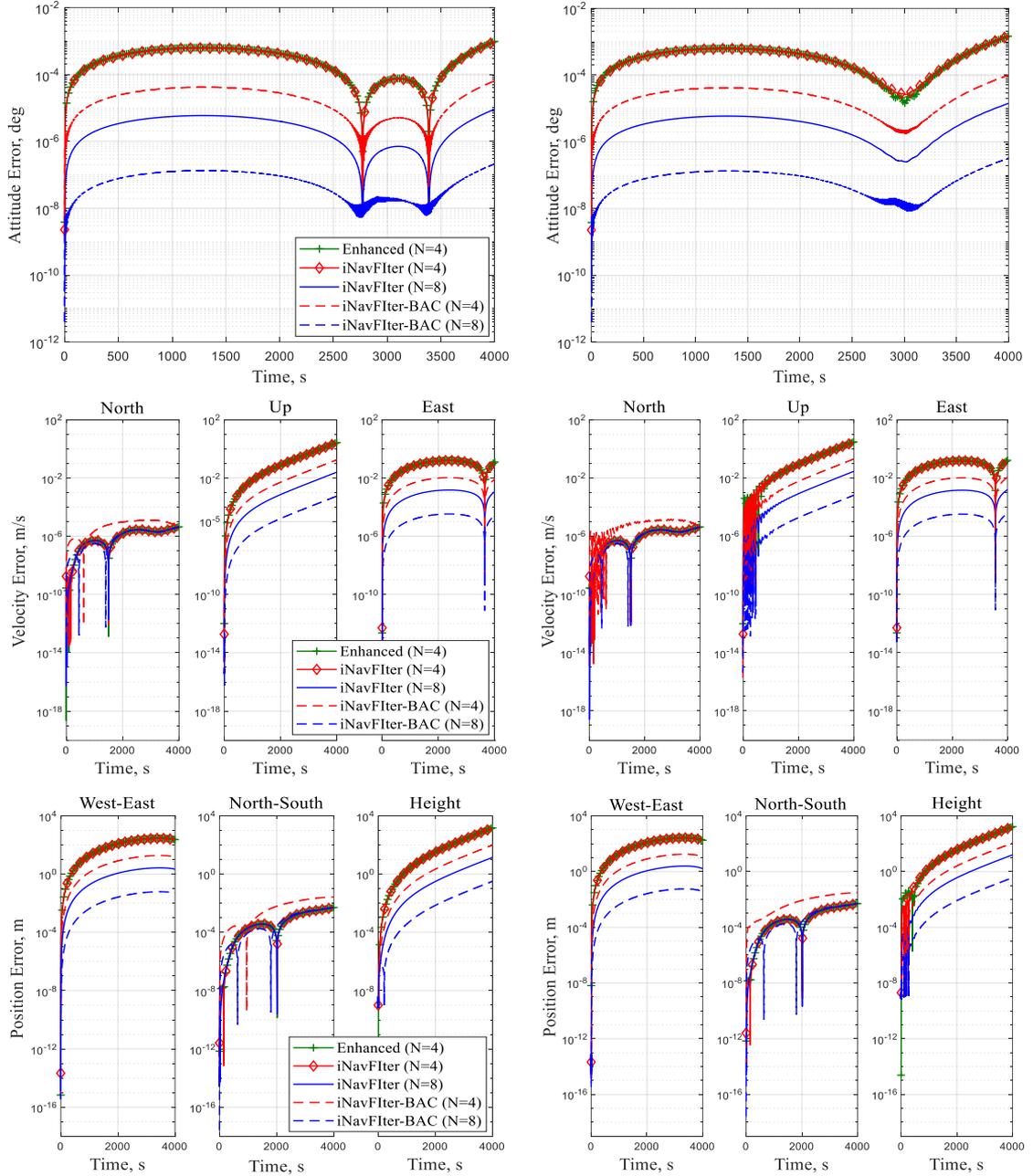

Fig. 3. Principal angle error, velocity and position errors for the enhanced algorithm and the functional iteration algorithms (left column for 1$g$ cruising condition, right column for non-cruising condition; 1/16 degree coning angle, 50 Hz coning input, 800 Hz sampling frequency; 'BAC' denotes the borrowing-and-cutting technique using extra samples in the neighboring computation windows.)

In addition, beyond the sustained large-angle coning environment, we also delve into a generalized vibratory environment, focusing on comparing the enhanced four-sample algorithm against the functional iteration four-sample algorithms, separately under a 1$g$ cruising condition and a non-cruising condition. Specifically, the coning frequency is set as 50 Hz to mimic a vibratory environment, and the coning angle is chosen as 1/16 degree. Under the 1$g$ cruising condition, the vehicle moves eastward at a constant speed of $v = 500 \, m/s$, while the velocity variation under the non-cruising condition aligns with the parameter setting used in the aforementioned entire inertial navigation computation scenario. Taking 800 Hz sampling frequency as an example to illustrate, Fig. 3 presents the error curves under 1$g$ cruising and non-cruising conditions, respectively. The results show that at the 800 Hz sampling frequency, the enhanced four-sample algorithm and the functional iteration four-sample algorithm exhibit roughly the same accuracy in both conditions. However, more samples like N=8 help improve the overall computing accuracy of the functional iteration algorithms by about two orders, as a result of better approximation of angular velocity/specific force inputs for functional iteration. Furthermore, the borrowing-and-cutting (BAC) technique in [32] for refining the angular velocity/specific force inputs (by using extra samples in the neighboring computation windows) also contributes remarkably to the accuracy of both iNavFIter (N=4) and iNavFIter (N=8). Reversely, this observation indicates that it is the insufficient polynomial approximation of angular velocity/specific force inputs by using N=4 samples that leads to the roughly same accuracy of iNavFIter (N=4) and Enhanced (N=4), in whether 1$g$ cruising or non-cruising conditions.

## V. CONCLUSION

This paper investigates and compares the accuracy performance of the traditional and enhanced algorithms against the functional iteration algorithms. With the help of MATLAB Symbolic Toolbox, the analyses indicate that the enhanced attitude algorithm notably reduces error orders of the traditional counterpart, while the impact of the enhanced velocity algorithm on error order reduction is insignificant. Simulation results indicate that the traditional/enhanced two-sample algorithms have the same accuracy performance in both scenarios of the body-frame attitude computation and the entire inertial navigation computation, while the enhanced four-sample algorithm notably improves accuracy in the body-frame attitude scenario, but this superiority substantially diminishes in the entire inertial navigation computation scenario. This phenomenon is caused by the insufficiency of the traditional and enhanced velocity/position computation, especially the inadequate compensation of reference-frame rotation effect. It somewhat throws lights on the adverse effect of historically putting less efforts on velocity/position (and its coupling with attitude) research than those on attitude research. Meanwhile, the functional iteration algorithm still has a significant accuracy superiority under sustained lowly dynamic conditions. For the highly dynamic motion, it is imperative to appropriately increase the number of samples to improve the computation accuracy of any algorithm, as reported in the previous literature.

## ACKNOWLEDGMENT

The authors sincerely thank Dr. Yury A. Litmanovich for his valuable and constructive suggestions.